\documentclass[10pt,twocolumn,letterpaper]{article}

\usepackage{iccv}
\usepackage{times}
\usepackage{epsfig}
\usepackage{graphicx}
\usepackage{amsmath}
\usepackage{amssymb}

\usepackage{mathptmx}
\usepackage[scaled=0.92]{helvet}
\usepackage{microtype}
\usepackage{lipsum}

\newcommand\blfootnote[1]{%
  \begingroup
  \renewcommand\thefootnote{}\footnote{#1}%
  \addtocounter{footnote}{-1}%
  \endgroup
}

\usepackage[breaklinks=true,bookmarks=false]{hyperref}

\iccvfinalcopy % *** Uncomment this line for the final submission

\ificcvfinal\fi

\begin{document}

\title{FaceEraser: Removing Facial Parts for Augmented Reality}

\author{Miao Hua* \qquad Lijie Liu* \qquad Ziyang Cheng* \qquad Qian He \qquad Bingchuan Li \qquad Zili Yi† \\
Bytedance Inc.\\
{\tt\small \{huamiao,liulijie.gxz,chengziyang.yy,heqian,libingchuan,yizili\}@bytedance.com}
}

\maketitle
\ificcvfinal\thispagestyle{empty}\fi

\begin{abstract}
\blfootnote{*C0-first author; †Corresponding author}
Our task is to remove all facial parts (e.g., eyebrows, eyes, mouth and nose), and then impose visual elements onto the ``blank'' face for augmented reality.  Conventional object removal methods rely on image inpainting techniques (e.g., EdgeConnect\cite{nazeri2019edgeconnect}, HiFill\cite{yi2020contextual}) that are trained in a self-supervised manner with randomly manipulated image pairs. Specifically, given a set of natural images, randomly masked images are used as inputs and the raw images are treated as ground truths. Whereas, this technique does not satisfy the requirements of facial parts removal, as it is hard to obtain ``ground-truth'' images with real ``blank'' faces. Simple techniques such as color averaging or PatchMatch \cite{barnes2009patchmatch} fail to assure texture or color coherency. To address this issue, we propose a novel data generation technique to produce paired training data that well mimic the ``blank'' faces. In the mean time, we propose a novel network architecture for improved inpainting quality for our task. Finally, we demonstrate various face-oriented augmented reality applications on top of our facial parts removal model. The source codes are released at \href{https://github.com/duxingren14/FaceEraser}{duxingren14/FaceEraser} on github for research purposes.
\end{abstract}

\section{Introduction}

\begin{figure}[t]
\includegraphics[width=\linewidth]{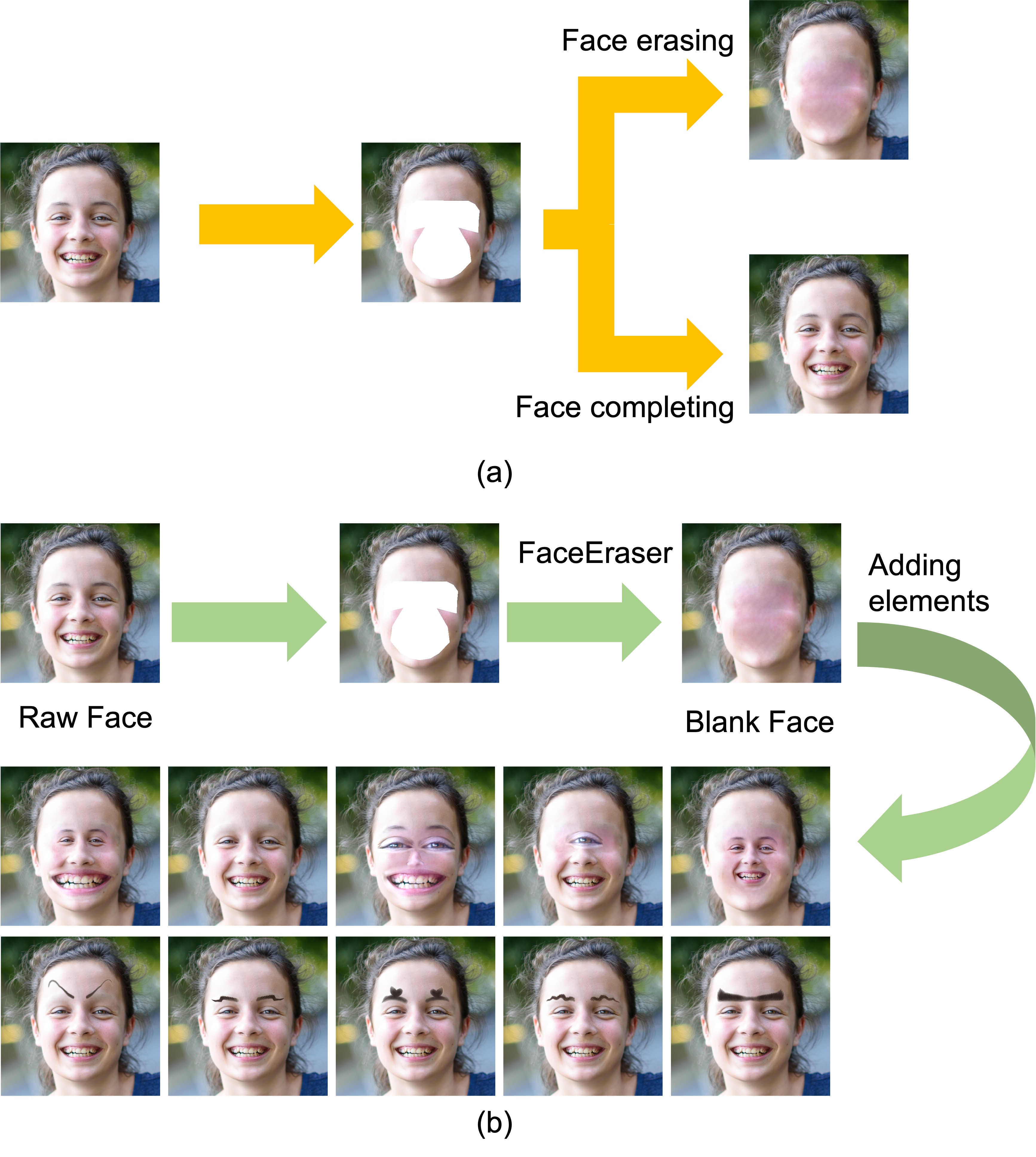}
\centering
\caption{\textbf{(a)} The difference between face completing and face erasing. Note that existing models are mostly trained to complete a face rather than erase a face. \textbf{(b)} Exemplar applications of face-oriented augmented reality that require face erasing. \textbf{Please zoom in for higher resolution.}}
\label{fig:teaser}
\end{figure}

Augmented reality applications are becoming increasingly popular among users due to the possibility of combining real-world elements with virtual ones. Specifically, face-oriented augmented reality is widely used in pictures and videos on social media platforms. In some applications the face is augmented by directly placing virtual elements (e.g., cat whisker, sun glasses, virtual eyebrows or makeups) onto it, while some other applications require facial parts (e.g., eyebrows, nose, mouth) to be removed before imposing virtual ones: see Figure \ref{fig:teaser} (b). In this paper, we are focusing on automatic facial parts removal for added value to face-oriented augmented reality applications. 

We propose to remove facial parts and obtain a ``blank'' face. However, existing image inpainting methods \cite{nazeri2019edgeconnect,yi2020contextual,yu2019free,liu2018image} attempt to auto-complete facial parts rather than removing them, which is limited to the conventional way that generates training pairs by randomly masking real faces and learns to complete masked pixels: see Figure~\ref{fig:teaser} (a). A more elaborate investigation implies that there is a lack of research into problems of this type. 

As one of the earliest studies into this problem, we propose to solve the problem of facial parts removal using a novel data preparation technique. We still follow the paradigm of image inpainting models, i.e., given a facial image and a mask indicating the segments of facial parts to be removed, a model is trained to generate a ``blank'' face with all facial parts removed. The novel data preparation method enables us to produce a number of images that well simulate the ``blank'''  faces we want. With these ``blank'' faces, we can then generate training pairs by randomly masking the images. In addition to the data preparation method, we also elaborate a novel network architecture specifically for the task of facial parts removal which exhibits improved quality over existing inpainting models. With the inpainting model trained, we can then process facial parts removal upon real faces, and demonstrate various interesting augmented reality applications by superposing manipulated or virtual elements onto the ``blank'' faces. 

The contributions of this paper include:
 
\begin{itemize}
\item{we propose a novel method to prepare the training data for the task of facial parts removal.}
\item{we design a novel network architecture that introduce the mechanism similar to patch-copying-based inpainting methods\cite{barnes2009patchmatch,criminisi2004region,criminisi2003object}, which we call ``pixel-clone'', into neural inpainting models for improved inpainting quality in our task, effectively resolving the color or texture inconsistency issue observed in existing inpainting methods (e.g., EdgeConnect \cite{nazeri2019edgeconnect} and StructureFlow \cite{ren2019structureflow}).}
\item{We implement and demonstrate various augmented reality applications enabled by our facial parts removal model.}
\end{itemize}

\section{Related work}
\subsection{Image inpainting}

Image inpainting has been an active research area for the past few decades. Existing methods for image hole filling are either rule-based (e.g., Patch-based~\cite{criminisi2004region,criminisi2003object}, diffusion-based~\cite{bertalmio2000image,ballester2001filling}) or learning-based (Contextual Attention~\cite{yu2018generative,yu2019free}, HiFill \cite{yi2020contextual}, Partial Convolution~\cite{liu2018image}, EdgeConnect~\cite{nazeri2019edgeconnect}, StructureFlow~\cite{ren2019structureflow}).  

Rule-based methods~\cite{criminisi2004region,criminisi2003object,bertalmio2000image,ballester2001filling} attempt to fill the hole by extending and borrowing real patches from surrounding regions, whereas they lack awareness of the semantic structure of contents. Learning-based methods \cite{yu2018generative,yu2019free,liu2018image,yi2020contextual,nazeri2019edgeconnect} hallucinate missing pixels in a data-driven manner with the use of large external databases. These approaches learn to model the distribution of the training images and assume that regions surrounded by similar contexts likely to possess similar contents. These methods are more content-aware, but sometimes suffer inconsistency of colors or lack of fine-grained textures. Some researchers attempt to combine the two streams of research, explicitly introducing the ``patch copying'' mechanism into neural networks. These approaches include Contextual Attention \cite{yu2018generative,yu2019free,yi2020contextual}, StructureFlow \cite{ren2019structureflow}, Patch-Swap layer \cite{song2018contextual}. Specifically, Yu et al.~\cite{yu2018generative} introduce a novel contextual attention layer that enables borrowing features from distant spatial locations. Song et al. \cite{song2018contextual} employ the patch-swap layer that propagates the high-frequency texture details from the boundaries to hole regions. Ren et al. \cite{ren2019structureflow} introduce the novel appearance flow to warp features from surroundings to the hole region with the expectation to yield better image details. 

Similarly, we adopt the ``patch copying'' idea in our model. Whereas, different from these methods that copy patches from outside-hole regions at the feature level, which is prone to exert ``grid patterns'' or ``boundary artifacts'', we are copying at the pixel level. Unlike Yu et al.~\cite{yu2018generative} that soft-blend multiple nearest neighbors, we only copy one nearest neighbor, which turns out to generate sharper results.

\subsection{Facial analysis}

We use a face detection model to discover faces from wild images. 106 facial key points \cite{liu2019grand} are detected and used to align faces, locate facial parts and generate facial masks, which facilitates the procedure of data preparation, inference and augmented reality applications. We use facial attributes analyzers to predict the existence of glasses and hats. We employ glasses segmentation and hair segmentation models to acquire masks of glasses and hair.

\begin{figure*}[t]
\includegraphics[width=\linewidth]{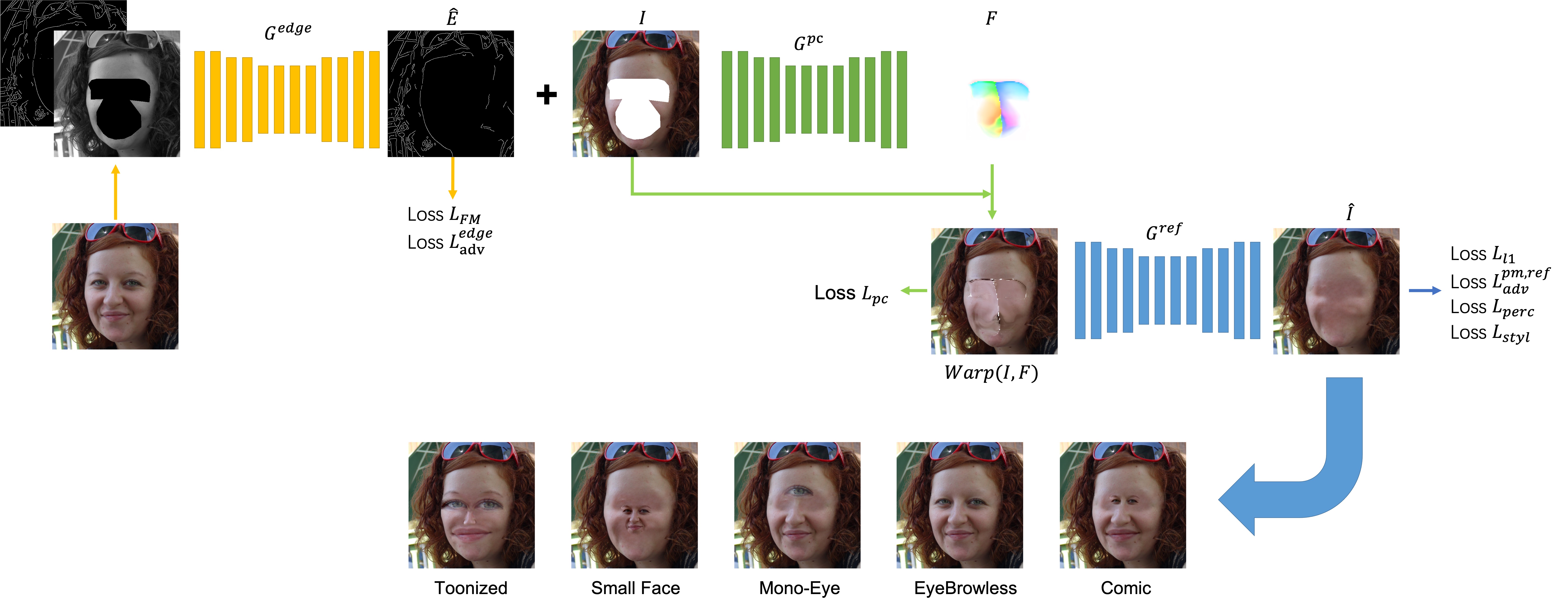}
\centering
\caption{The overall process of our face erasing model .}
\label{fig:process}
\end{figure*}

\section{Method}

\subsection{Overall pipeline}

We propose an image inpainting model consisting of three sub-networks: Edge Completion Network $G^{edge}$, Pixel-Clone Network $G^{pc}$ and Refine Network $G^{ref}$ (Figure \ref{fig:process}). The Edge Completion Network is an edge generator that hallucinates edges of the missing region of the image, the Pixel-Clone Network predicts the nearest neighbors of missing pixels from the visible contents with the guidance of hallucinated edges, and the Refine Network generates the final result using the nearest neighbors as a priori. With such design, we expect the hallucinated edges by $G_{edge}$ can provide some hints to the ``pixel-cloning'' mechanism of  $G^{pc}$ so that pixels be copied from the correct region.

Specifically, given an incomplete input image $\textbf{I}$ and a mask $\textbf{M}$ with value 1 indicating the hole region and 0 indicating the visible region. Pixel values of the input image are scaled to the range of $[0,1]$ before being processed. The Edge Completion Network takes the grayscale version of the input image $\textbf{I}^{gray}$, mask $M$ and the masked edge map $\textbf{E}$ as inputs, and generates a completed edge map $\textbf{\^{E}}=\textbf{G}^{edge}(\textbf{I}^{gray}, \textbf{E}, \textbf{M})$. The masked edge map here is extracted from $\textbf{I}^{gray}$ using canny algorithm with Gaussian width of $\sigma=2.0$, low threshold of 10\% and high threshold of 20\% \cite{nazeri2019edgeconnect}. 

The Pixel-Clone Network $\textbf{G}^{pc}$ that searches the nearest neighbor from the visible region of $\textbf{I}$ for each in-hole pixels. The output of Pixel-Clone Network is a two-channel flow map of the same size of $\textbf{I}$, indicating the x-y offsets of the nearest neighbor of each pixel, $\textbf{F}=\textbf{G}^{pc}(\textbf{\^{E}}, \textbf{I}, \textbf{M})$. With the predicted offsets  $\textbf{F}$, we conduct image warping upon $\textbf{I}$ at the image-level and obtain a coarse inpainting result, $\textbf{Warp}(\textbf{I},\textbf{F})$. Finally, the Refine Network $\textbf{G}^{ref}$ refines from the coarse result and generates $\textbf{\^{I}} = \textbf{G}^{ref}(\textbf{Warp}(\textbf{I}, \textbf{F}), \textbf{I}, \textbf{F})$.

%The architecture of three sub-networks are elaborated in Section \ref{sect:architecture}. The training method of the model are provided in Section \ref{sect:loss}.

\subsection{Data preparation}
\label{sect:data}
\paragraph{Training image preparation} 

\begin{figure}[t]
\includegraphics[width=\linewidth]{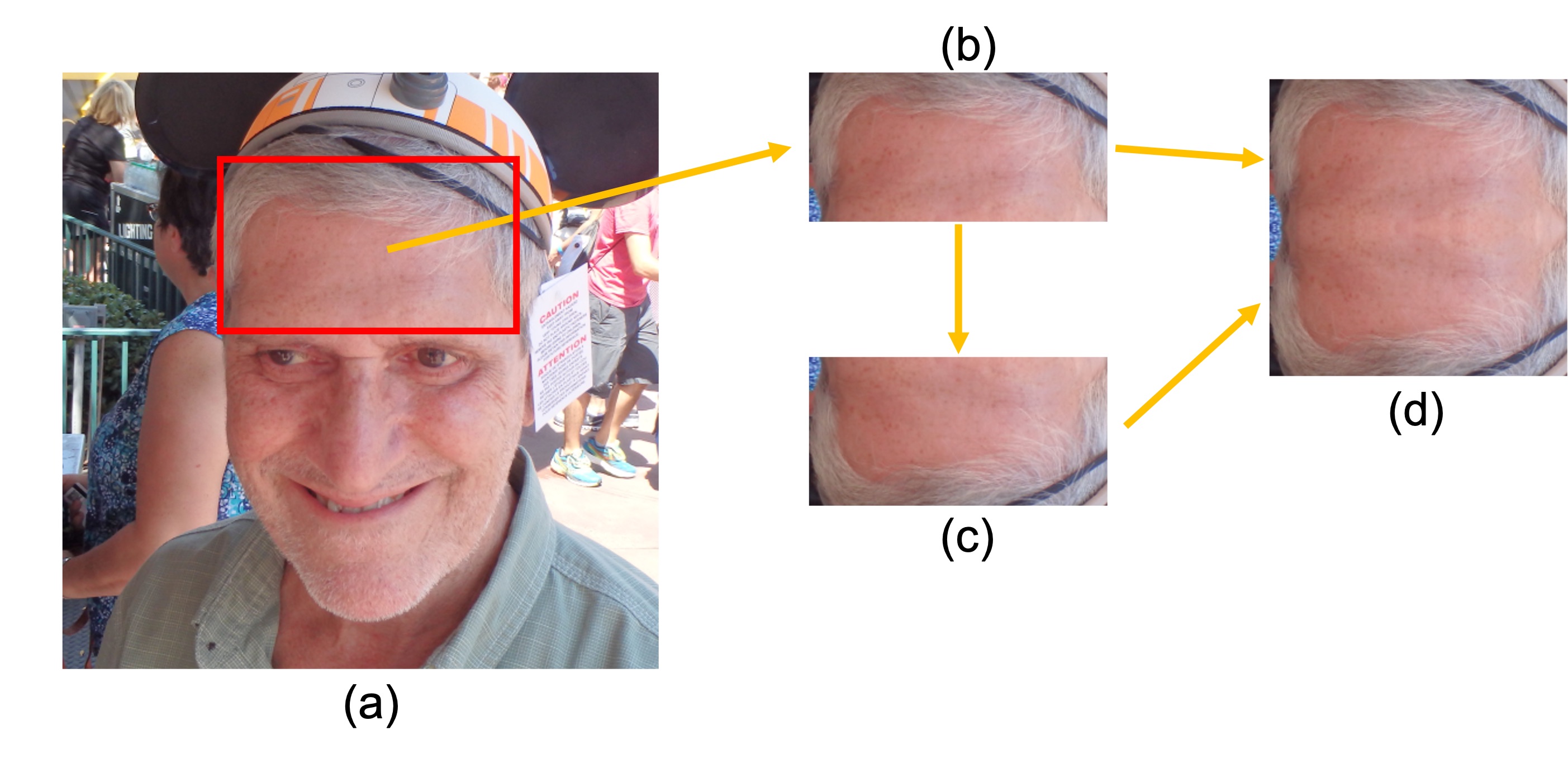}
\centering
\caption{The procedure of training image preparation. \textbf{(a)} raw facial image. \textbf{(b)} forehead image cropped from (a) based on key points. \textbf{(c)} vertical flip of (b). \textbf{(d)} stitching (b) and (c)}
\label{fig:data}
\end{figure}

Our task is to inpaint a masked face with skin textures and obtain a ``blank'' face: see Figure \ref{fig:teaser}. As there are no real ``blank'' faces in the world, to make the training of image inpainting models possible, we need to generate a number of ``blank'' faces that well simulate what we desire. The procedure of training image preparation is illustrated in Figure \ref{fig:data}.

Given a set of real face images, we remove images coming with glasses and hats, and those whose foreheads are largely covered by hair. After the filtering, we detect the key points of the face, and crop the forehead region of each face above the eyebrow. We then resize the forehead image to 256$\times$128. Finally, we vertically flip the forehead image and stitch it with the original forehead image to acquire a 256$\times$256 image that looks like a ``blank'' face: see Figure \ref{fig:data}.

\paragraph{Mask preparation}

\begin{figure}[t]
\includegraphics[width=\linewidth]{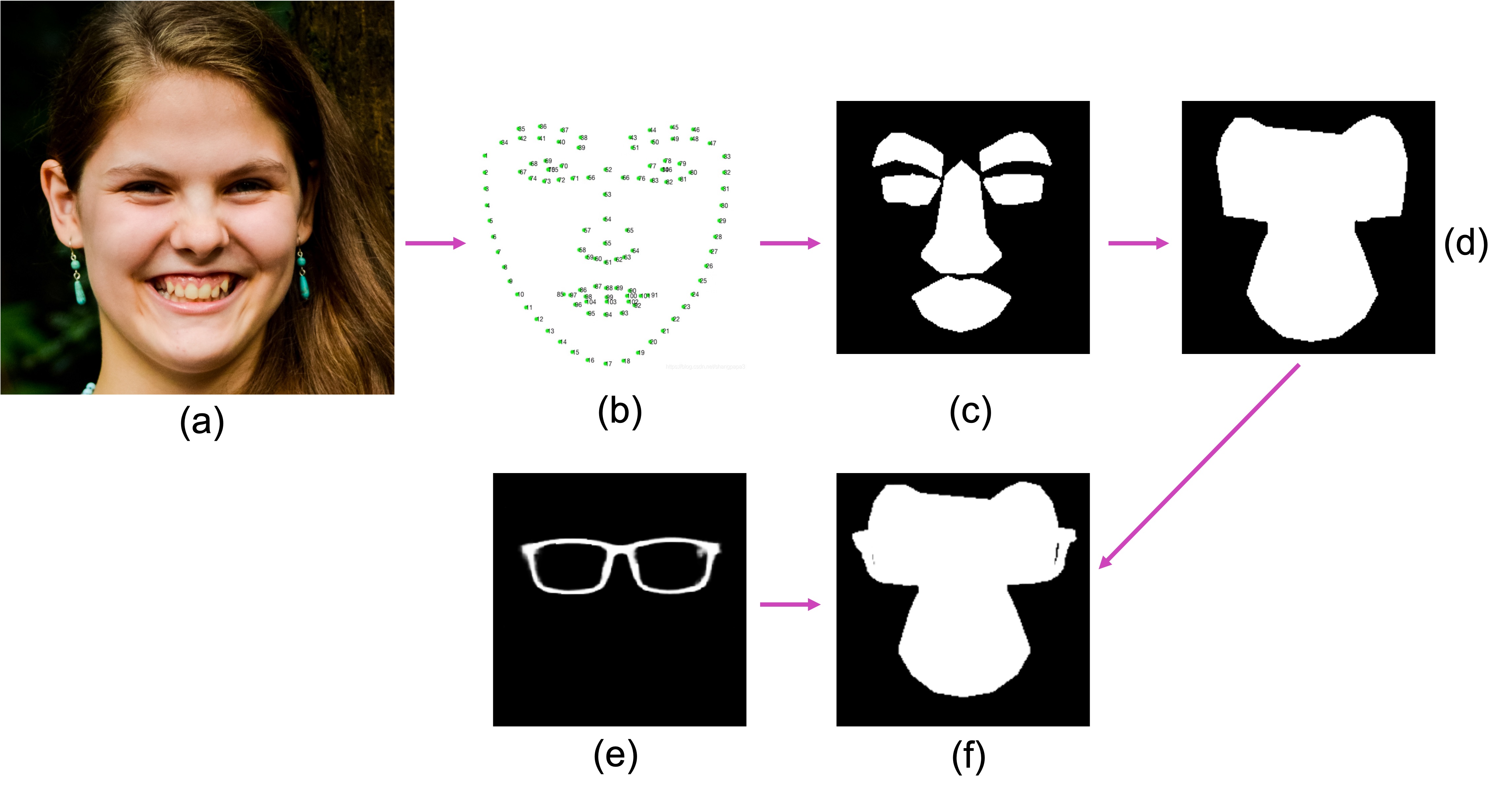}
\centering
\caption{The procedure of mask generation. \textbf{(a)} raw image. \textbf{(b)} 106 key points detected from (a). \textbf{(c)} initial mask obtained by connecting key points in (b). \textbf{(d)} mask dilated from (c). \textbf{(e)} real glasses mask estimated with glasses segmentation model. \textbf{(f)} merging (e) and (d).}
\label{fig:mask}
\end{figure}

The masks we used for training should serve the task of facial part removal. We acquire the training masks from real faces. The procedure is shown in Figure \ref{fig:mask}. We first extract the 106 key points from pre-aligned faces. We then infer the masks of facial parts by connecting the key points. Finally, we dilate the masks to assure covering of shadows and peripheral regions. In some cases, the existence of glasses should also be taken into account. Therefore, we augment a fraction of masks with glasses masks. The glasses masks are obtained from real images with an internal glasses segmentation model.

%We make use of face landmarks estimated with our internal API, and then the part masks are inferred from the landmarks by connecting the key points to form a polygon enclosing the a region. After that, hair mask is gained with a public available hair segmentation model. 

\subsection{Network architecture}
\label{sect:architecture}

Our model is comprised of three sub-networks, Edge Completion Network $\textbf{G}^{edge}$, Pixel-Clone Network $\textbf{G}^{pc}$, and Refine Network $\textbf{G}^{ref}$.  
The Edge Completion Network $\textbf{G}^{edge}$ follows an encoder-decoder architecture similar to the method proposed by Nazeri et al. \cite{nazeri2019edgeconnect}, which proves to be effective in connecting corrupted edges. Specifically, the generator consists of encoders that down-sample twice, followed by eight residual blocks \cite{nazeri2019edgeconnect} and decoders that upsample images back to the original size. Dilated convolutions \cite{yu2015multi} with a dilation factor of two are used in the residual layers. Instance normalization and ReLU are used for all convolutions except the last convolution which is activated with \textit{Sigmoid} to scale pixel values to range of $(0,1)$.

The Pixel-Clone Network $G^{pc}$ follows almost identical encoder-decoder architecture as the Edge Completion Network except that the last activation layer is replaced with \textit{tanh} to scale output values to $(-1,1)$. The number of input and output channels also differ. $G^{pc}$ predicts a dense offset map $\textbf{F}$, indicating the x-y offsets of the nearest neighbor of each pixel. Specifically, it is a two-channel map of the same size of the input image. 

% where 1 denotes displacement of half of the image width (or height) along the positive direction of  the x-axis (or y-axis) while -1 denotes displacement of half width (or height) along the negative direction

The coarse inpainting result is obtained by warping the input image $\textbf{I}$ with offset map $\textbf{F}$, i.e.,  $\textbf{\^{I}}=\textbf{Warp}(\textbf{I}, \textbf{F})$. This can be done with the  \textit{grid\_sample} operation in PyTorch, which is differentiable over both inputs.
 
We utilize a U-Net architecture for the Refine Network $\textbf{G}^{ref}$. The Refine Network is composed of 5 downsampling residual blocks followed 5 upsampling residual blocks. The skip connections are joined through addition instead of concatenation, which proves to be more memory-efficient.  We use channel attention \cite{zhang2018image} in each residual block, as it proves to add improvements for low-level tasks such as super-resolution, as it adaptively rescales channel-wise features by considering interdependencies among channels. We verify that the introduction of channel attention drives improvements to the inpainting quality. 

 \paragraph{Discriminators}
We need two discriminators for adversarial training of the generator, respectively for the training of the Edge Completion Network and the joint training of Patch-Clone Network and Refine Network. For both discriminators,  we use the same PatchGAN \cite{isola2017image,yi2017dualgan} architecture that rates the realness of 70$\times$70 patches \cite{isola2017image,nazeri2019edgeconnect}  rather than the entire image. We use spectral normalization \cite{miyato2018spectral} on discriminators to stabilize the training by scaling down weight matrices by their respective largest singular values. We use leaky ReLU as activations except for the last layer that uses Sigmoid.

\subsection{Training method}
\label{sect:loss}

Given the incomplete input image $\textbf{I}$, mask $\textbf{M}$ and the ground-truth image $\textbf{I}_{gt}$, the Edge Completion Network predicts a completed edge map $\textbf{\^{E}}=\textbf{G}^{edge}(\textbf{I}^{gray}, \textbf{E}, \textbf{M})$, where $\textbf{I}^{gray}$ is the grayscale version of the input image $\textbf{I}$ and $\textbf{E}$ is the incomplete edge map of $\textbf{I}^{gray}$.  Similar to \cite{nazeri2019edgeconnect}, we employ a two-phase training procedure. In Phase 1, the Edge Completion Network is trained independently. After the full convergence of the Edge Completion Network $\textbf{G}^{edge}$, we then train the Pixel-Clone Network and Refine Network jointly while holding the weights of $\textbf{G}^{edge}$ fixed. We did not find joint training or finetuning of all three sub-networks brings any significant quality improvements.

The Edge Completion Network is trained with an objective comprised of an adversarial loss \cite{goodfellow2014generative} and feature-matching loss \cite{wang2018high}, which is written as

\begin{equation}
 \label{eq:edge}
 \begin{split}
\min_{G^{edge}} \max_{D^{edge}} L^{edge} =  \min_{G^{edge}}\alpha_{adv}^{edge} \max_{D^{edge}}L_{adv} +  \alpha_{FM} L_{FM}
\end{split}
\end{equation}

where $\alpha_{adv}^{edge}$ and $\alpha_{FM}$ are the coefficients of the adversarial loss $L_{adv}$ and feature matching loss $L_{FM}$ respectively. The adversarial loss $L_{adv}$ is defined as

\begin{equation}
\label{eq:edge_adv}
\begin{split}
 L^{edge}_{adv} = \mathop{{}\mathbb{E}}_{I_{gt}, M} \left[ \log D^{edge}(E_{gt}, I_{gt}^{gray})  + \log (1 - D^{edge}( \hat{E}, I_{gt}^{gray})) \right] 
\end{split}
\end{equation}

where $\textbf{I}_{gt}^{gray}$ and $\textbf{E}_{gt}$ are the grayscale version and the edge map of the ground truth image $\textbf{I}_{gt}$. The feature-matching loss $L_{FM}$ compares the activations of the fake and the real edge maps in the intermediate layers of the discriminator, and forces the generator to produce results with representations that are similar to the real edge maps. The feature matching loss is written as

 \begin{equation}
\label{eq:edge_fm}
\begin{split}
 L_{FM} = \mathop{{}\mathbb{E}}_{\textbf{I}_{gt}, M} \left[\sum_{i=1}^L \frac{1}{N_i} \left\| D^{edge}_i (\hat{E}, I_{gt}^{gray})- D^{edge}_i (E_{gt},I_{gt}^{gray}) \right\|_1 \right]
\end{split}
\end{equation}

where $L$ is the number of convolutional layers of the discriminator, $D^{edge}_i$ is the output of the \textit{i }-th activation layer of the discriminator and $N_i$ is the number of elements in the \textit{i}-th activation layer. We use $L1$ norm to encourage less blurring \cite{isola2017image}. In our experiments, we choose $\alpha_{adv}^{edge}=1$ and $\alpha_{FM}=10.$

 %=============================== inpainting  network

The Pixel-Clone Network $\textbf{G}^{pc}$ that searches the pixels from the visible region to fill in-hole pixels, is one of our key contributions. The output of Pixel-Clone Network is a two-channel offset map indicating the x-y offsets of the nearest neighbor of each pixel, which is denoted as $\textbf{F}=\textbf{G}^{pc}(\textbf{\^{E}}, \textbf{I}, \textbf{M})$. The coarse inpainting result is obtained by warping the input, i.e. $\textbf{Warp}(\textbf{I},\textbf{F})$. The Refine Network $\textbf{G}^{ref}$ generate a refined result from  the coarse, namely $\textbf{\^{I}} = \textbf{G}^{ref}(\textbf{Warp}(\textbf{I}, \textbf{F}), \textbf{I}, \textbf{F})$.

The Pixel-Clone Network $\textbf{G}^{pc}$ and Refine Network $\textbf{G}^{ref}$ are trained jointly with loss $L^{pc,ref} $, which is defined as

 \begin{equation}
 \label{eq:ref}
 \begin{split}
 L^{pc,ref} =  \alpha_{adv} L^{pc,ref}_{adv} +  \alpha_{perc} L_{perc} +  \alpha_{L1} L_{L1} +  \alpha_{styl} L_{styl}  +  \alpha_{pc} L_{pc} 
\end{split}
\end{equation}

where $L^{pc,ref}_{adv}$, $L_{perc}$, $L_{L1}$, $L_{styl}$ and $L_{pc}$ are the adversarial loss, perceptual loss, L1 loss, style loss and Pixel-Clone loss term respectively, and $\alpha_{adv}$,  $\alpha_{perc}$,  $\alpha_{L1}$,  $\alpha_{styl}$ and  $\alpha_{pc}$ are the corresponding coefficients. The adversarial loss is defined as

\begin{equation}
\label{eq:ref_adv}
\begin{split}
 L^{pc,ref}_{adv} = \mathop{{}\mathbb{E}}_{I_{gt}, M} \left[log D^{pc,ref}( I_{gt}, \hat{E}) +   
 log (1 - D^{pc,ref}(\hat{I}, \hat{E})) \right] 
\end{split}
\end{equation}

We adopt the perceptual loss $L_{perc}$ and style loss $L_{styl}$ proposed in \cite{johnson2016perceptual} . $L_{perc}$ penalizes the perceptual disparities of the fake to the ground-truth by measuring the distance between the activations of a pre-trained network, which is defined as

 \begin{equation}
\label{eq:perc}
\begin{split}
 L_{perc} = \mathop{{}\mathbb{E}}_{\textbf{I}_{gt}, M}   \left[\sum_{i=1}^L \frac{1}{N_i} \left\| \phi_i (\hat{I})-  \phi_i (I_{gt}) \right\|_1 \right]
\end{split}
\end{equation}

where $\phi_i$ is the output of \textit{i}-th activation layer of a pre-trained network. In our method,  $\phi_i$ corresponds to the output of activation layers $relu1_1$, $relu2_1$, $relu3_1$, $relu4_1$ and $relu5_1$ of the VGG-19 network \cite{simonyan2014very} pre-trained on the ImageNet dataset \cite{russakovsky2015imagenet}. These activation maps are also used to compute style loss which is defined as

 \begin{equation}
\label{eq:styl}
\begin{split}
 L_{styl} = \mathop{{}\mathbb{E}} _{\textbf{I}_{gt}, M}  \left[\sum_{i=1}^L  \left\| G^{\phi}_i (\hat{I})-  G^{\phi}_i (I_{gt}) \right\|_1 \right ]
\end{split}
\end{equation}

where $G^{\phi}_i (\cdot)$ is the $C_i \times C_i$ Gram matrix \cite{johnson2016perceptual} computed from \textit{i}-th activation maps $\phi_i$. Style loss measures the differences between covariance of the features.

 The $L1$ loss and Pixel-Clone loss $L_{pc}$ are computed as

 \begin{equation}
\label{eq:l1}
\begin{split}
 L_{L1} = \mathop{{}\mathbb{E}}_{\textbf{I}_{gt}, M}  \left\| \hat{I}-  I_{gt} \right\|_1
\end{split}
\end{equation}

 \begin{equation}
\label{eq:pc}
\begin{split}
 L_{pc} = \mathop{{}\mathbb{E}}_{\textbf{I}_{gt}, M}  \left\| WarP(I, F)-  I_{gt} \right\|_1
\end{split}
\end{equation}

We choose $\alpha_{adv}=0.1$,  $\alpha_{perc}=1.0$,  $\alpha_{L1}=1.0$,  $\alpha_{styl}=500$ and $\alpha_{pc}=1.0$ in our experiments.

\subsection{Face-oriented augmented reality}

On top of our face erasing model, we implement various face-oriented augmented reality applications including mono-eye, comic, small-face, toonized, eyebrowless effects: see Figure \ref{fig:teaser} (b), Figure \ref{fig:process}. For these effects, we follow the same procedure that all facial parts are removed first and then certain elements are pasted back to the ``blank'' face. Poison blending \cite{perez2003poisson} is used to avoid boundary artifacts when merging the elements with the face. Specifically, the comic effect involves enlarged mouth, shrunk eyes and proper adjustment of positions of the facial parts. As for eyebrowless effect, we simply paste back all the facial parts except for the eyebrows. As for the toonized effect, we paste back the eyebrows, enlarged eyes and mouth, shrunk nose and properly adjust the positions of facial parts. As for the small face effect, we simply shrink the face and paste it back. As for the mono-eye effect, we paste back the mouth and nose, and place one eye right above the nose.

In addition to the applications mentioned above, the erased faces enable various augmented-reality applications that require placement of any user-customized elements.

\begin{figure*}[t]
\includegraphics[width=.9\linewidth]{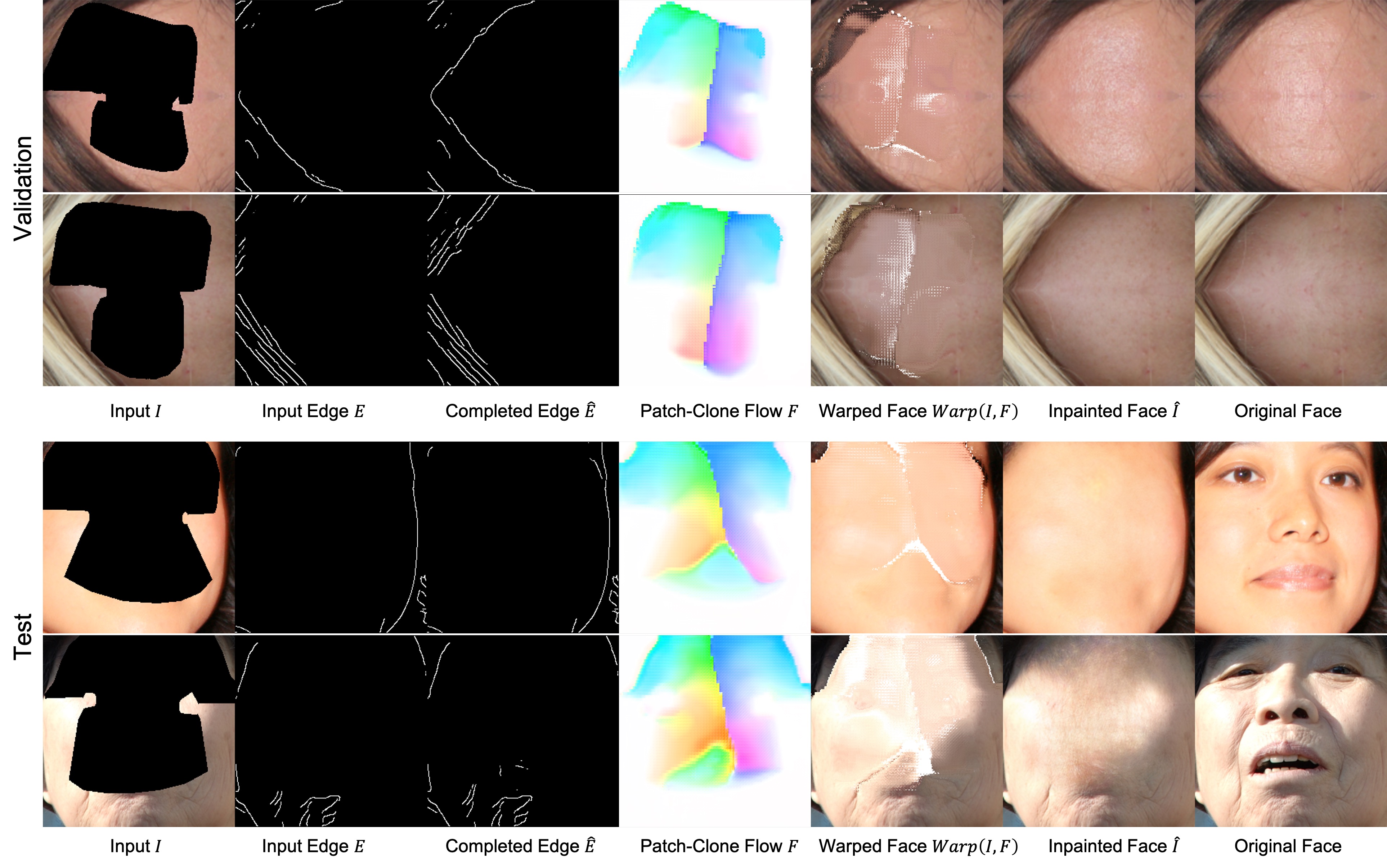}
\centering
\caption{Intermediate and final outputs of our method. Two uppers rows are results on the validation data (fake ``blank'' faces), and two lower rows are inference results obtained on real faces.}
\label{fig:example}
\end{figure*}

\section{Experimental results}

\subsection{Training setup  and strategies }
Training images and masks are obtained from FFHQ dataset~\cite{karras2019style} by following the procedure as described in Section \ref{sect:data}. We choose FFHQ dataset as it contains considerable variation in terms of age, ethnicity, poses and illumination conditions, and it has good coverage of accessories such as eyeglasses, sunglasses and hats. We obtain about 35,000 images and 10,000 masks for training, and 4000 images and 1000 masks for validation. Our model can do inference on  images from VoxCeleb \cite{nagrani2017voxceleb}, CelebA-HQ \cite{karras2017progressive}, unseen faces from FFHQ or any other unconstrained faces. The network is trained using 256 $\times$ 256 images with a batch size of 8. The model is optimized using Adam optimizer \cite{kingma2014adam} with $\beta_1$ = 0.0 and $\beta_2$ = 0.9. All generators are trained with learning rate $10^{-4}$, and discriminators are trained with learning rate of $10^{-5}$. 

The model is implemented in PyTorch. Each model is trained with one Tesla V100 GPU for 2000,000 epochs.

Figure \ref{fig:example} illustrates the intermediate and final results of our face erasing results on the validation and test splits. It can be observed that the completed edge map $\textbf{\^{E}}$ faithfully enforces the copying of pixels from the valid regions. The offset map $\textbf{F}$ accurately matches the nearest neighbor for a majority of ``in-hole'' pixels except for the pixels in the ``crack''. The Refine Network effectively eliminates the visible artifacts in the coarse results and fills the ``crack'' with plausible contents and textures.

\begin{figure*}[t]
\includegraphics[width=.9\linewidth]{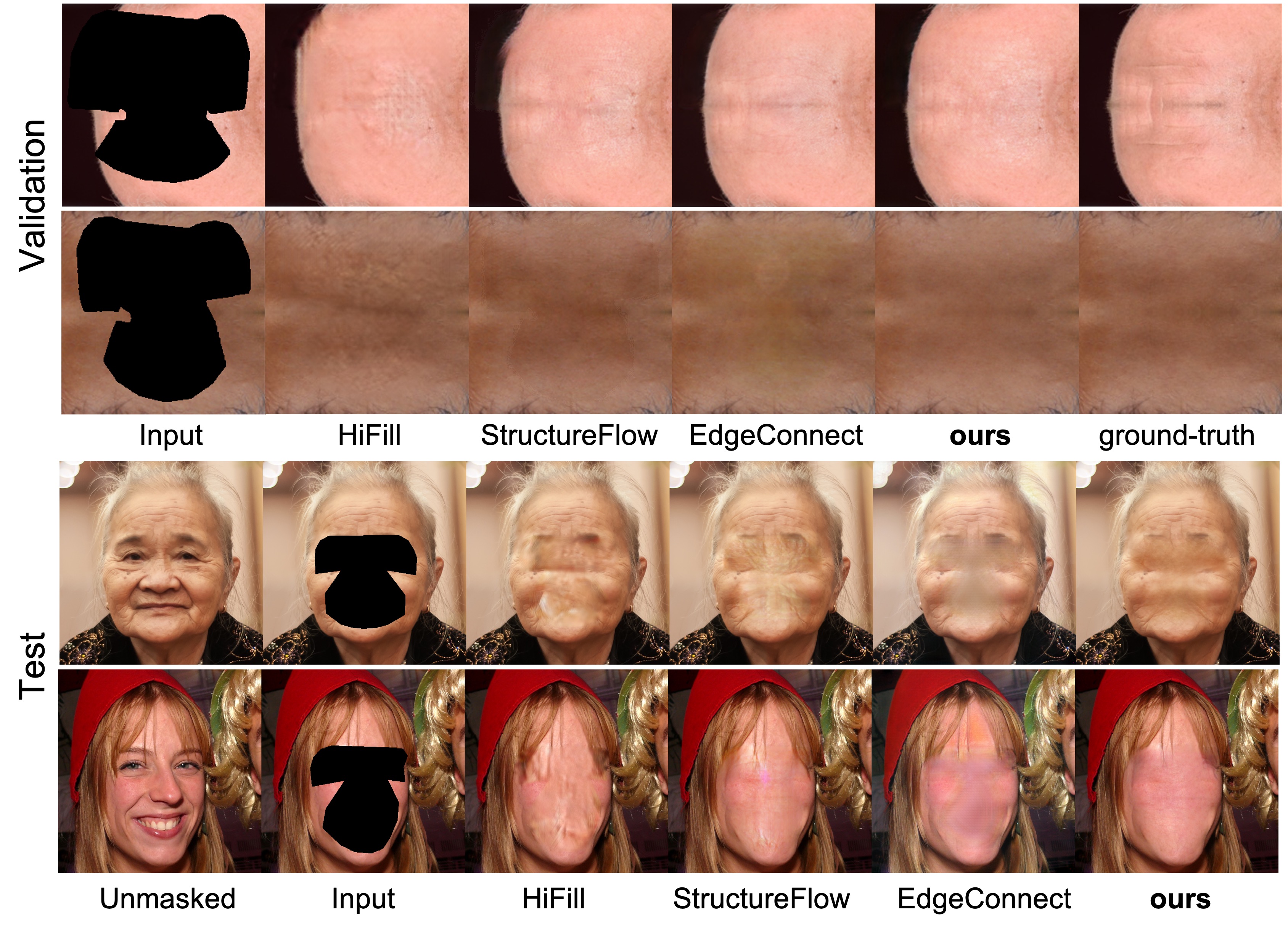}
\centering
\caption{Comparisons of our method with prior arts on the validation and test data. The upper two rows are the inpainting results on the validation data (fake ``blank'' faces produced as in Section \ref{sect:data}). The bottom two are testing results on real faces. }
\label{fig:comp1}
\end{figure*}

\subsection{Comparisons}

% \begin{center}
\begin{table}
 \caption{Quantitative evaluations on existing methods and different variations of our method. Here is the notations. Edge: Edge Completion Network \cite{nazeri2019edgeconnect}. PC: Pixel-Clone Network. UNet: Encoder-Decoder architecture with skip connections \cite{ronneberger2015u}. AE: Auto-Encoder without skip connections. CA: Channel Attention mechanism \cite{zhang2018image}.}
 \label{table:comp_ab}
 \resizebox{.5\textwidth}{!}{
\begin{tabular}{ l||c|c } 
  & PSNR & MAE   \\ 
   \hline
  \hline
 HiFill\cite{song2018contextual} & 28.24 $\pm$ 0.20 & 0.0253  $\pm$ 0.0035\\  
 StructureFlow\cite{ren2019structureflow} & 29.62 $\pm$ 0.23 & 0.0247  $\pm$ 0.0031\\  
   EdgeConnect (Edge+AE)\cite{nazeri2019edgeconnect} & 32.13 $\pm$ 0.25 & 0.0231  $\pm$ 0.0022\\  
 \textbf{Ours} (Edge+PC+AE) & 33.01 $\pm$ 0.25 & 0.0226  $\pm$ 0.0023\\  
 \textbf{Ours} (Edge+PC+UNet) & 33.11 $\pm$ 0.27 & 0.0209  $\pm$ 0.0019\\  
 \textbf{Ours} (Edge+PC+UNet+CA) & \textbf{33.17} $\pm$ 0.29 & \textbf{0.0199}  $\pm$ 0.0016\\  
 \end{tabular}}
\end{table}
%\end{center}

%(Edge+PM+UNet(DS)+CA) & \textbf{33.09} $\pm$ 0.28 & \textbf{0.0202}  $\pm$ 0.0019
% DS: Depth-Separable convolution \cite{howard2017mobilenets}

To verify the effectiveness of the proposed network architectures, we compared our method with the state-of-the-art image inpainting methods including HiFill \cite{song2018contextual}, StructureFlow \cite{ren2019structureflow}, EdgeConnect \cite{nazeri2019edgeconnect}. As the official pre-trained models are trained on either scene inpainting or facial completion tasks, which does not match our task, we retrained those models on the training datasets we produced.  To assure fair comparisons, all models are trained for 2000,000 epochs with the same batch size of 8. We attempt to use the same experimental setup for all these models though small disparities cannot be avoided. Specifically, HiFill only supports training and inferences on images no smaller than 512$\times$512. We resize all training samples to 512$\times$512 and do the training. As for inference, we resize the cropped face to  512$\times$512 to facilitate the inference and then resize it back after filling the hole. StructureFlow requires structure smoothing of the input images, we use python implementation of the structure filtering to replace the matlab implementation. For EdgeConnect which consists of an Edge Completion Network and Inpainting Network, we train the Edge Network for 1000,000 epochs and  jointly train the two networks for another 1000,000 epochs.

As shown in Figure \ref{fig:comp1}, we can observe that HiFill utilizes the contextual attention, soft-blending the nearest patches from visible regions, tends to generate blurry skin textures, and sometimes it mistakenly copy patches from the non-facial areas. StructureFlow can generate plausible contents to fill the holes. EdgeConnect generally outperforms HiFill and StructureFlow and generates plausible results. Though slight color and texture inconsistency can be observed. We also conducted quantitative evaluations of these methods on the validation dataset, in which we calculate the PSNR and mean absolute error (MAE) between the inpainted images and ground-truths.

\subsection{Ablation studies}
\label{sect:ablation}

 \begin{center}
\begin{table}
 \caption{Quantitative evaluations on the models trained with different $\alpha_{pc}$ values.}
 \label{table:ab1}
 \centering
\begin{tabular}{ c||c|c } 
  & PSNR & MAE   \\ 
   \hline
  \hline
$\alpha_{pc}=0.0$& 31.21 $\pm$ 0.25 & 0.0263  $\pm$ 0.0023\\  
$\alpha_{pc}=0.5$& 31.46 $\pm$ 0.29 & 0.0250  $\pm$ 0.0025\\  
$\alpha_{pc}=0.8$& 32.61 $\pm$ 0.30 & 0.0221  $\pm$ 0.0021\\  
$\alpha_{pc}=1.0$& \textbf{33.09} $\pm$ 0.28 & \textbf{0.0202}  $\pm$ 0.0019\\   
$\alpha_{pc}=1.2$& 32.87 $\pm$ 0.31 & 0.0205  $\pm$ 0.0019\\  
$\alpha_{pc}=1.5$& 32.56 $\pm$ 0.34 & 0.0209  $\pm$ 0.0020\\  
$\alpha_{pc}=2.0$& 32.38 $\pm$ 0.32 & 0.0214  $\pm$ 0.0017\\  
\end{tabular}
\end{table}
\end{center}

\begin{figure}[t]
\includegraphics[width=\linewidth]{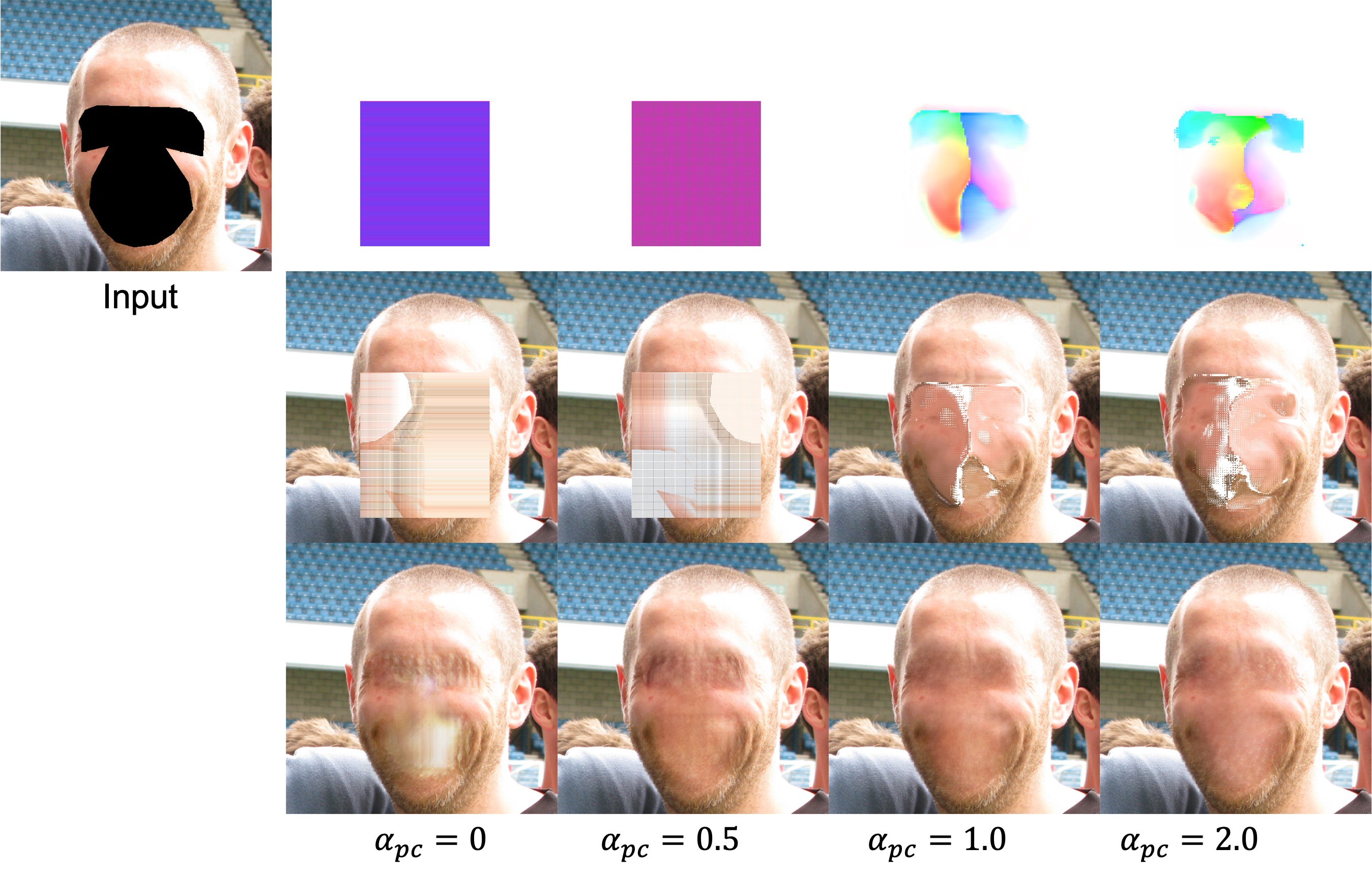}
\centering
\caption{Comparisons of inference results of models trained with different $\alpha_{pc}$ values by providing the same input as in the topleft. From top to bottom are $\textbf{F}$, $Warp(\textbf{I}, \textbf{F})$ and $\textbf{\^{I}}$ respectively.}
\label{fig:ab1}
\end{figure}

\begin{figure*}[t]
\includegraphics[width=0.8\linewidth]{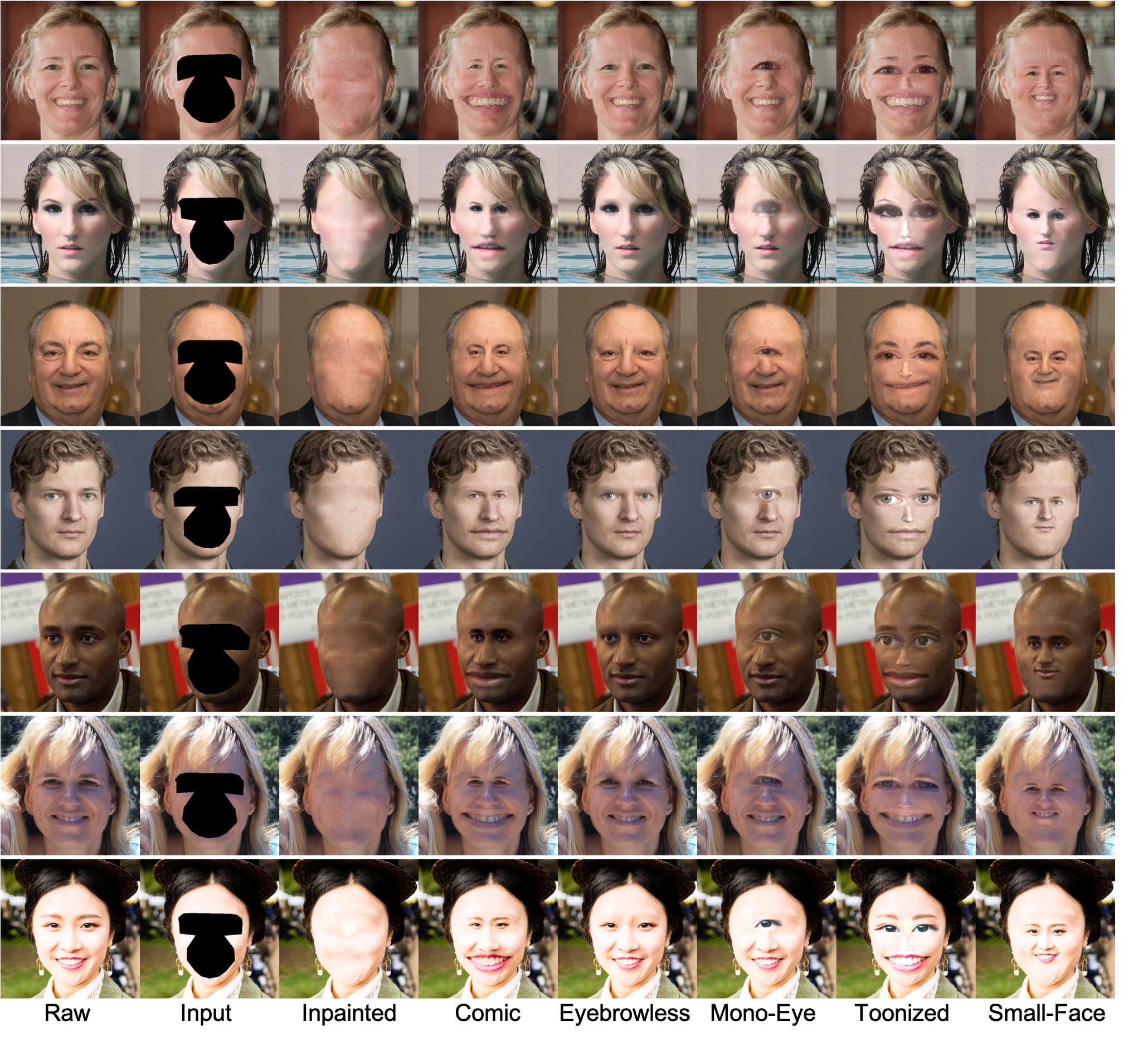}
\centering
\caption{Exemplar applications supported by our FaceEraser.}
\label{fig:ffhq}
\end{figure*}

We examine how Pixel-Clone loss term $L_{pc}$ affects the performance of our network.  Figure \ref{fig:ab1} shows that the inpainting results of models trained with different coefficient values $\alpha_{pc}$  for the Pixel-Clone loss term. As shown in Figure \ref{fig:ab1}, we observe that when $\alpha_{pc}<=0.5$, the offset map $\textbf{F}$ has constant value and does not effectively match the nearest neighbors, thus generating unsatisfactory inpainting results $\textbf{\^{I}}$. When $\alpha_{pc}=2.0$, the ``white dot'' artifacts become visible in $\textbf{\^{I}}$. The quantitative evaluations further verify the conclusion that $\alpha_{pc}=1.0$ is close to the optimal choice. 

Further, we did a quantitative evaluation of our methods in terms of different variations of the Refine Network. As the original EdgeConnect employs Edge Completion Network and Inpainting Network. We replace their Inpainting Network with a Pixel-Clone Network plus a Refine Network. In terms of the architecture of Refine Network, we experimented a few variations including AE and UNet. As shown in Table~\ref{table:comp_ab},  the UNet \cite{ronneberger2015u} architecture outperforms Auto-Encoder (AE) in our experimental setting, while Channel Attention (CA) \cite{zhang2018image} brings noticeable quality gain.

%The replacement of regular convolutions with Depth-Separable (DS) convolutions \cite{howard2017mobilenets} brings unnoticeable decrease in terms of quality but it is an order of magnitude faster.  

\subsection{Applications}

With the facial parts removed, we can then place a subset of manipulated facial parts or virtual elements on to the ``blank'' face to achieve interesting and enjoyable special effects.The augmented reality applications (e.g, mono-eye, small-face, eyebrowless, comic and toonized) are implemented and experimented on the test split of FFHQ dataset. The results are demonstrated in Figure \ref{fig:ffhq}. The test results on the CelebA-HQ \cite{karras2017progressive} and Vox-Celeb~\cite{nagrani2017voxceleb} datasets are provided in the supplementary materials. Further, we accelerate our model using techniques such as pruning, network slimming and quantization to allow real-time inference on mobile phones.   Some exemplar video demos are presented in the supplementary materials.

\section{Conclusion}
We propose a novel data preparation method and a novel network architecture for the task of facial parts removal (face erasing), a task that attracted insufficient attention from academia. The novel ``Pixel-Clone'' mechanism effectively solves the problem of the color and texture inconsistency issue observed in existing image inpainting models. We implement and demonstrate various interesting and enjoyable augmented-reality applications on top of our FaceEraser.

Our model suffers small chance of failure and is prone to generate unsatisfactory results due to undersegmentation of the facial parts (e.g., eyebrows, glasses) or interference of beards, hair or peripherals (see the supplementary materials for more details).

{\small
\bibliographystyle{ieee_fullname}
\bibliography{references}
}

\clearpage

\section{Appendix}
\begin{figure*}[t]
\includegraphics[width=\linewidth]{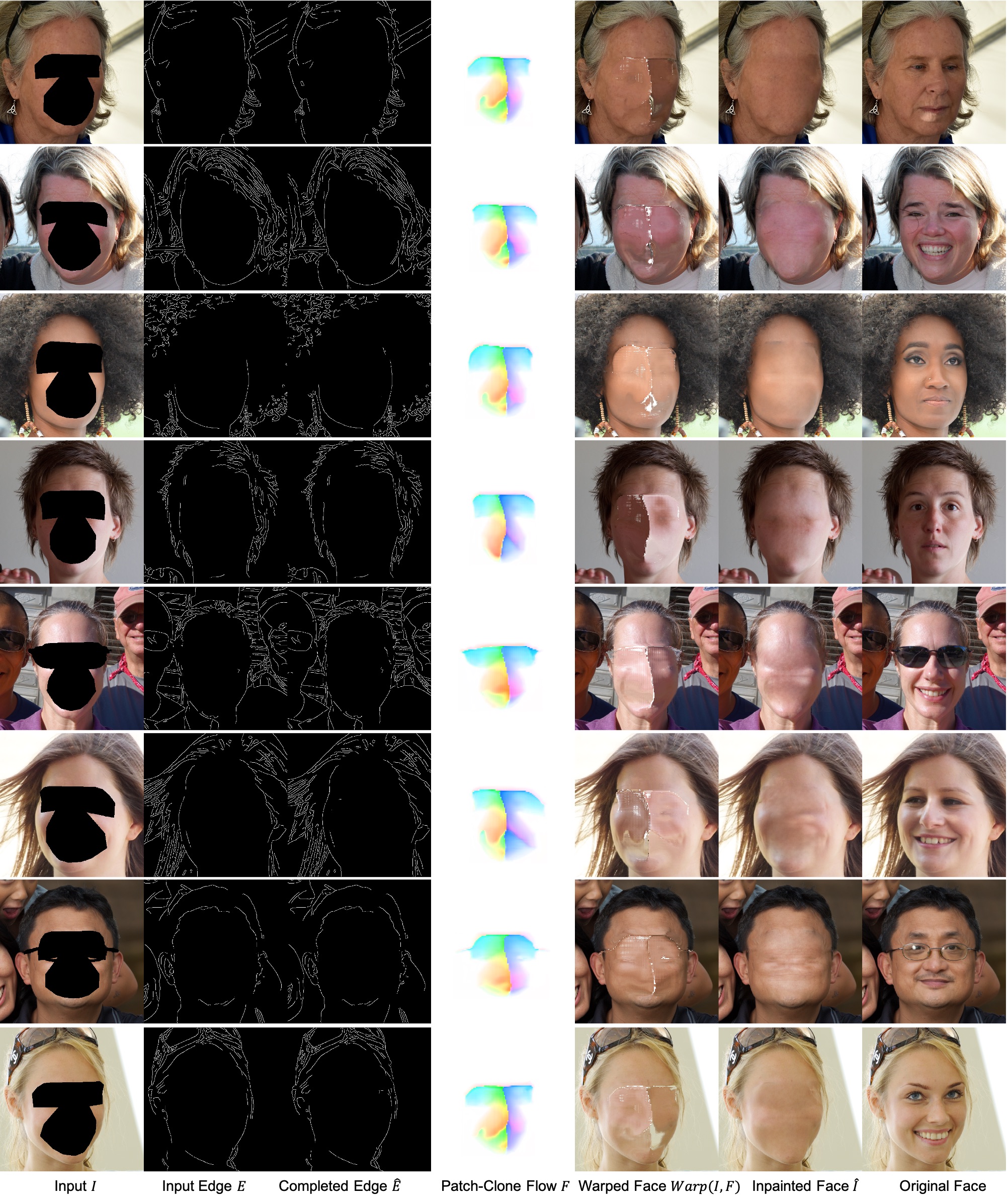}
\centering
\caption{Illustration of intermediate and final results of our inpainting models. The test images cones from the FFHQ test split, which are unseen by the pre-trained model.}
\label{fig:example2}
\end{figure*}

\begin{figure*}[t]
\includegraphics[width=\linewidth]{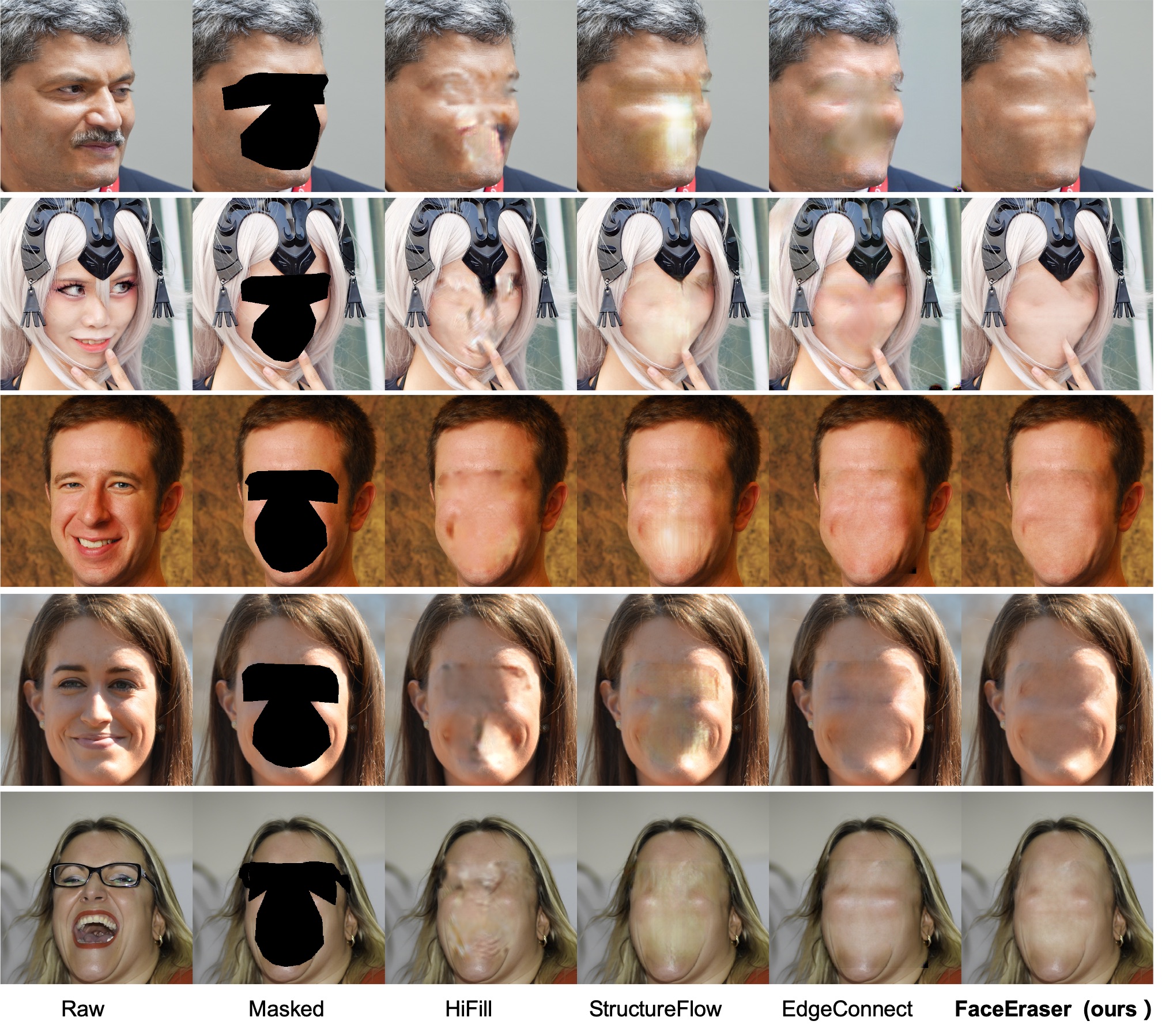}
\centering
\caption{Visual comparisons of our method with state-of-the-art inpainting methods. Note that the competing models are re-trained on our training datasets with similar configurations.}
\label{fig:comp2}
\end{figure*}

\begin{figure*}[t]
\includegraphics[width=\linewidth]{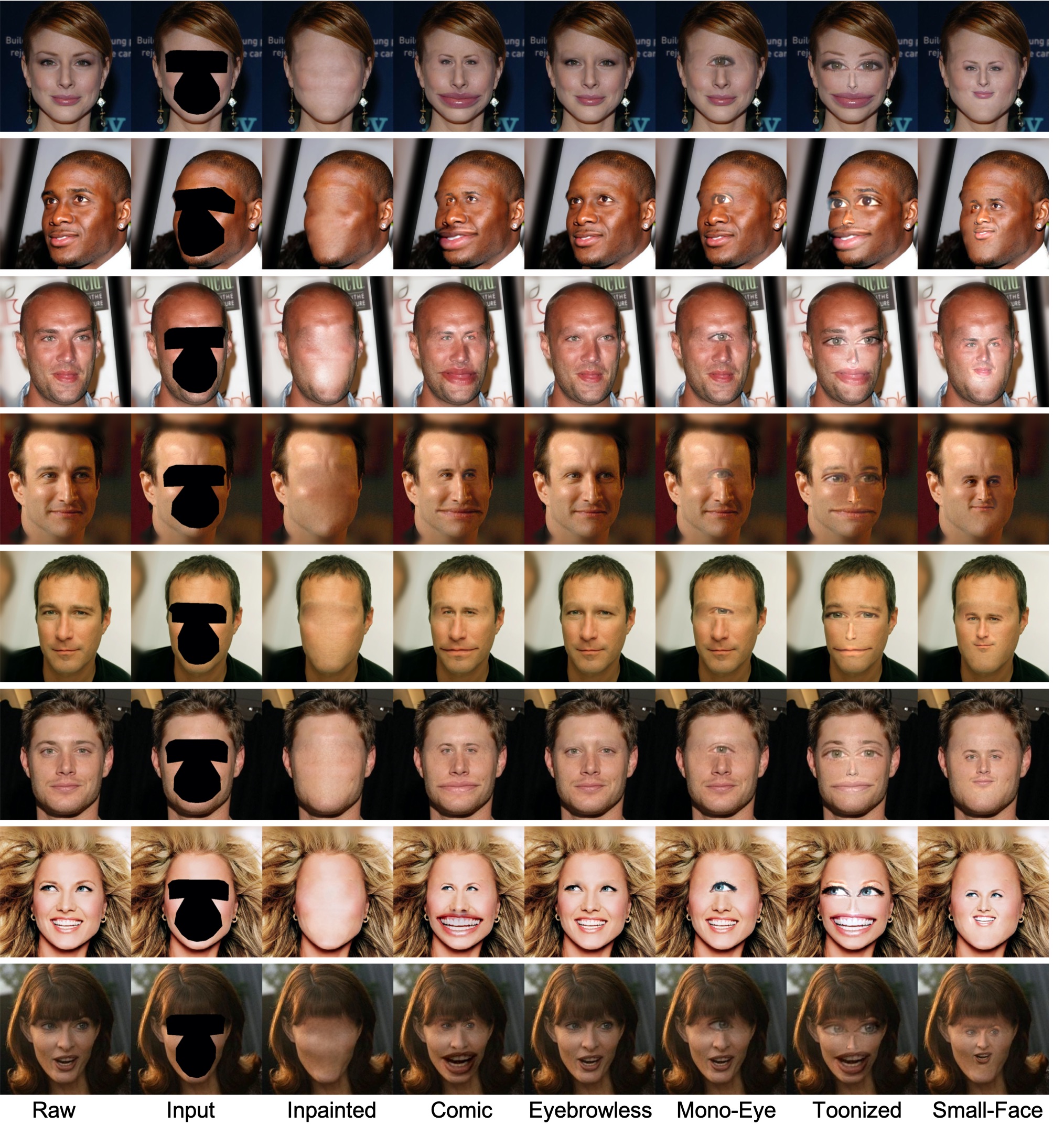}
\centering
\caption{Augmented-reality applications experimented on the CelebA-HQ dataset.}
\label{fig:celebahq}
\end{figure*}

\begin{figure*}[t]
\includegraphics[width=\linewidth]{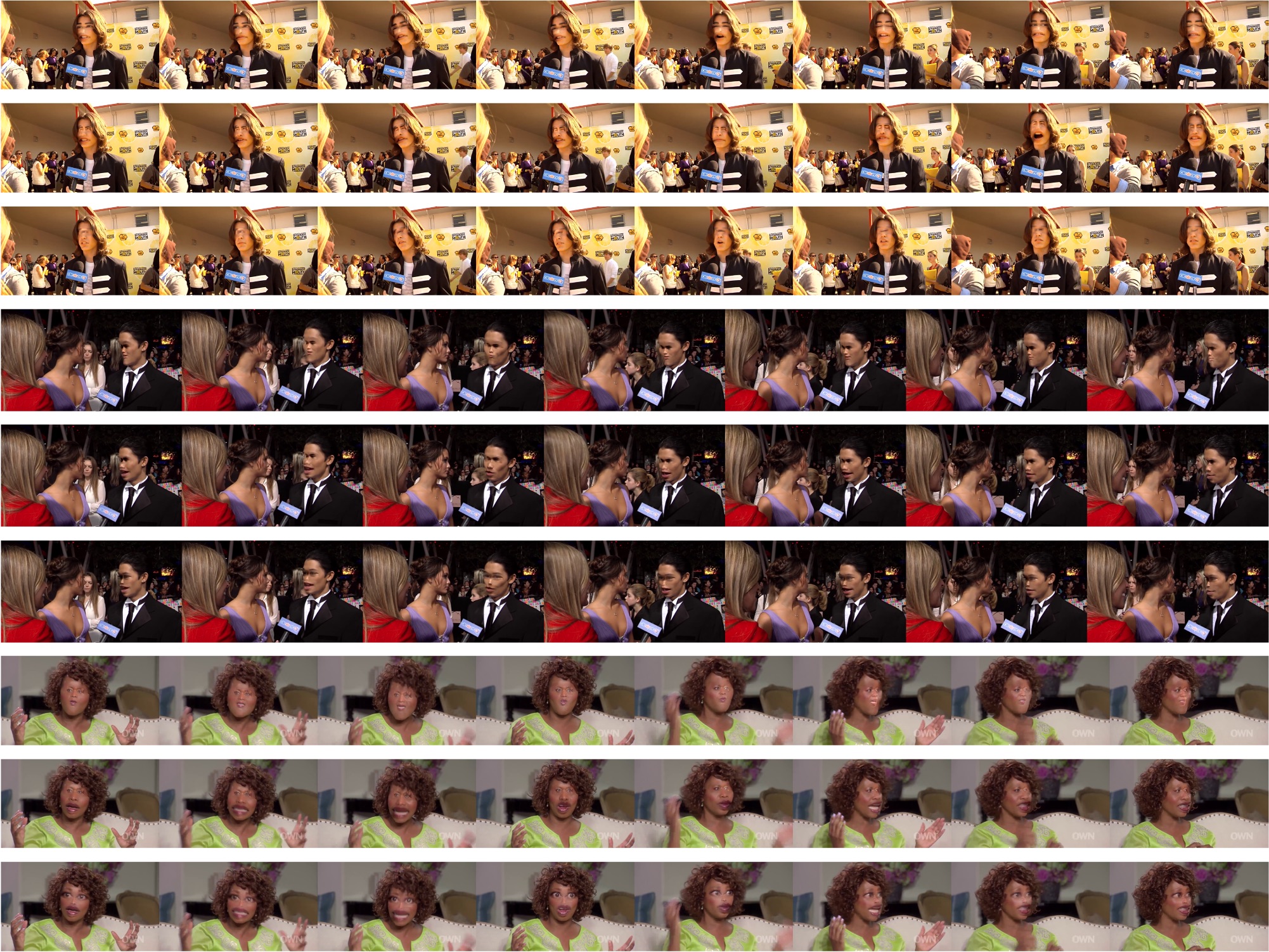}
\centering
\caption{Augmented-reality applications experimented on the VoxCeleb dataset.}
\label{fig:vox}
\end{figure*}

We present more intermediate and final results of our method on FFHQ test split in Figure \ref{fig:example2}.  
The demo results on CelebA-HQ\cite{karras2017progressive} and more VoxCeleb \cite{nagrani2017voxceleb} videos are presented in Figure \ref{fig:celebahq} and Figure \ref{fig:vox}.
We also present more comparative results of our method and existing methods in Figure \ref{fig:comp2}.

Some failure examples are presented in Figure \ref{fig:failure}.

Some more video demos are provided in the supplementary materials.
%Some of the training logs are presented in Figure .

\begin{figure*}[t]
\includegraphics[width=\linewidth]{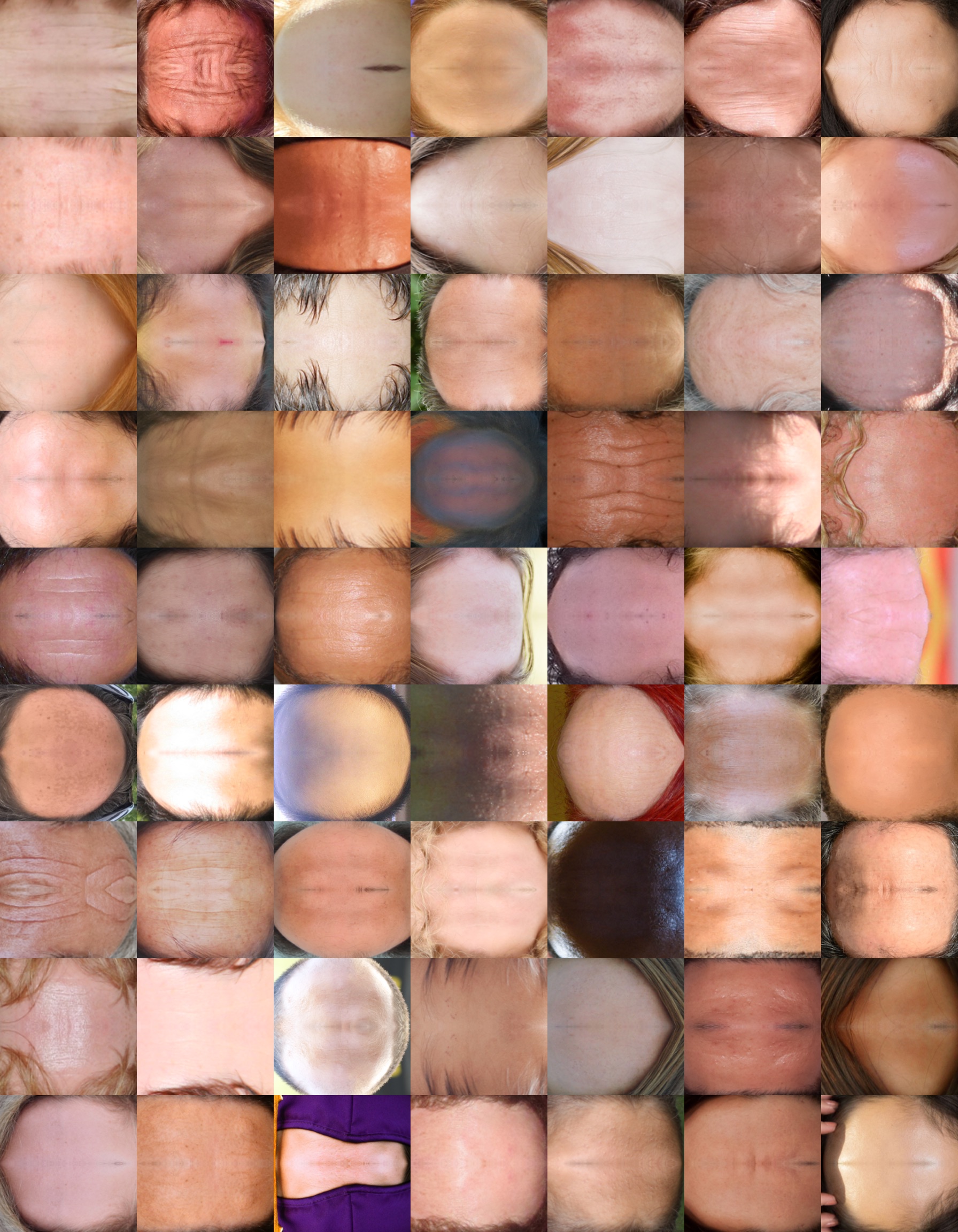}
\centering
\caption{Examples of training samples.}
\label{fig:train_imgs}
\end{figure*}

\begin{figure*}[t]
\includegraphics[width=\linewidth]{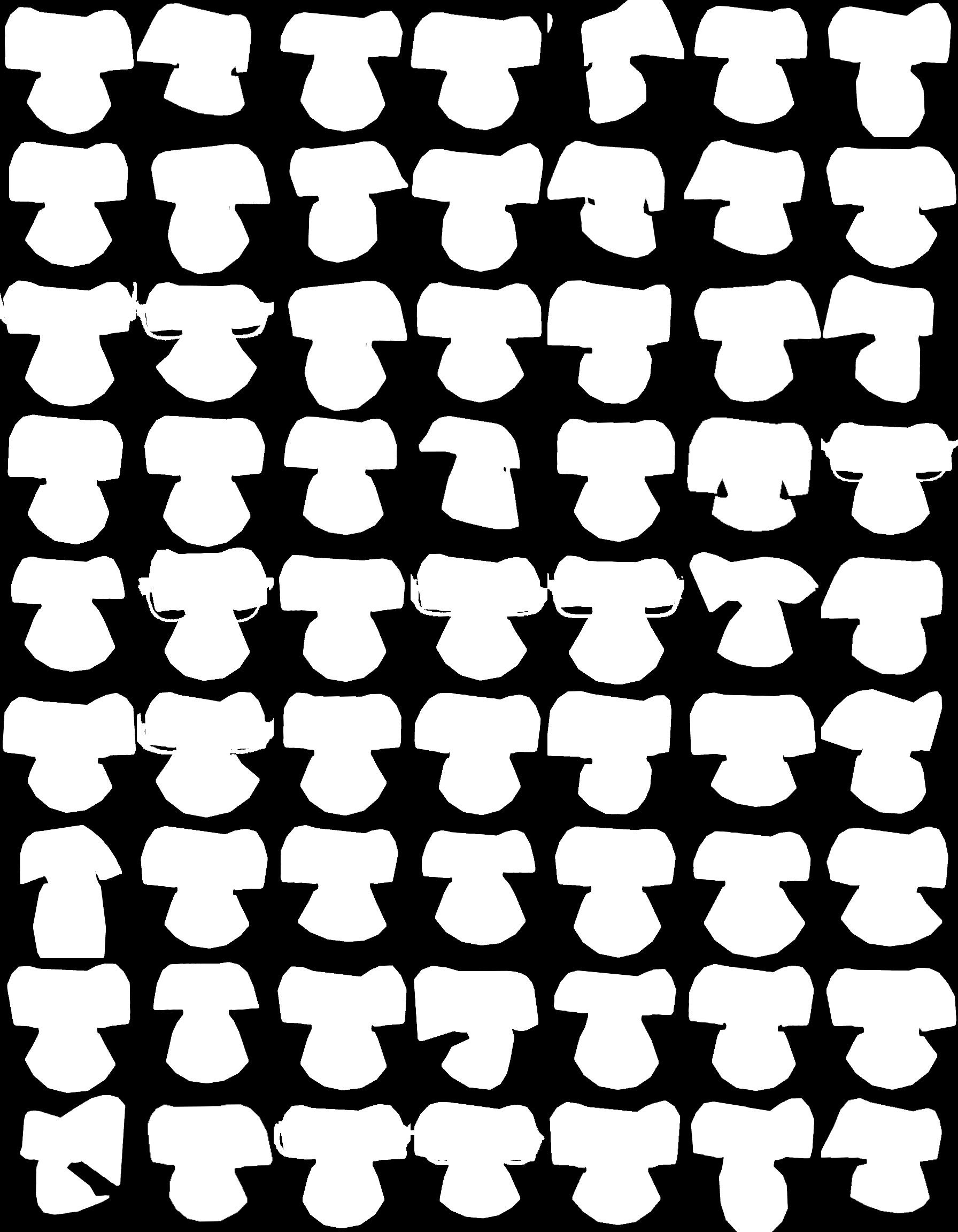}
\centering
\caption{Examples of training masks.}
\label{fig:train_masks}
\end{figure*}

\begin{figure*}[t]
\includegraphics[width=.9\linewidth]{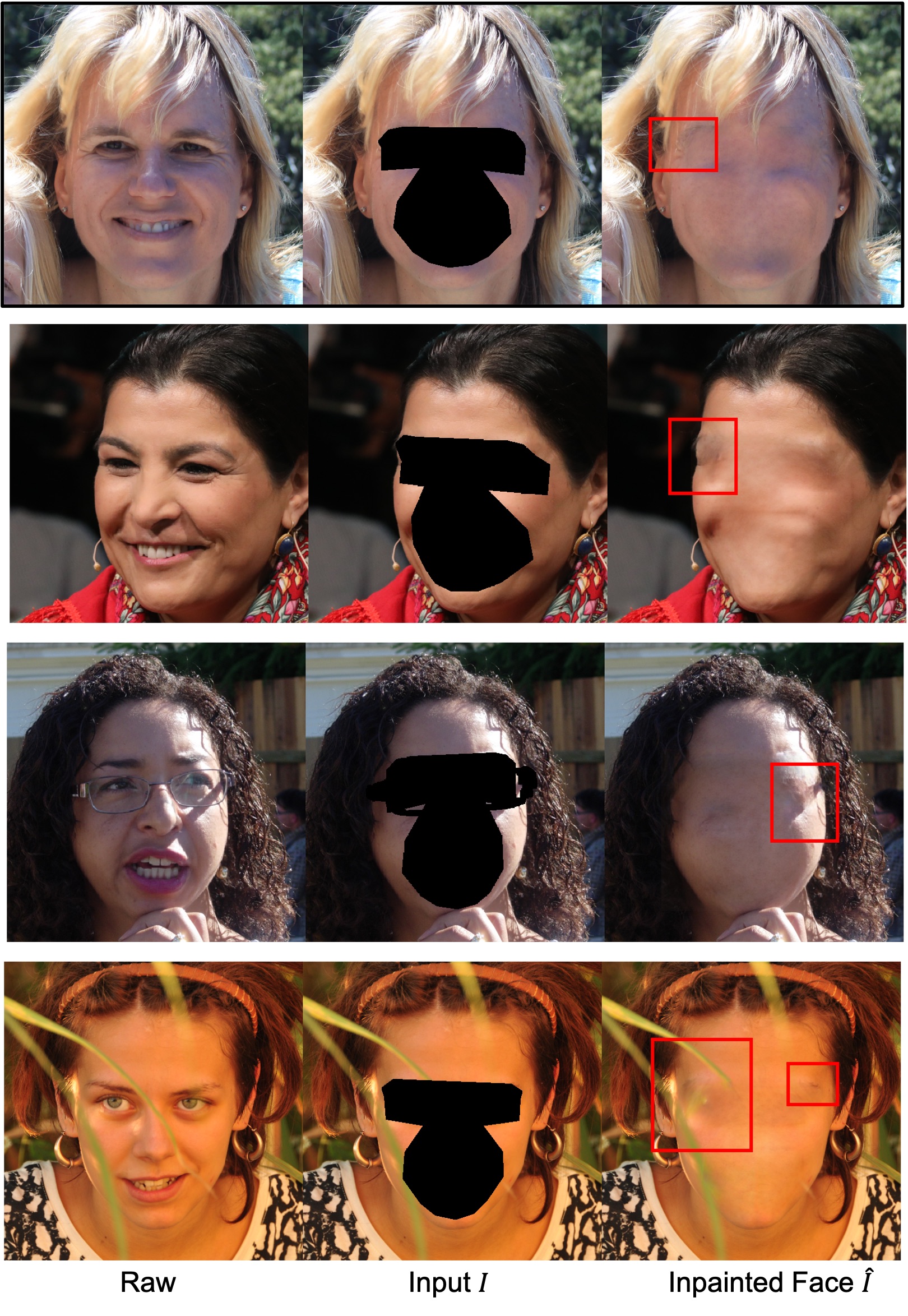}
\centering
\caption{Exemplar failure samples generated by our FaceEraser. The failures typically result from interference of hair, shadow or other peripherals, or undersegmentation of eyebrows or beards. The red boxes highlight the unpleasing artifacts.}
\label{fig:failure}
\end{figure*}

\end{document}